\documentclass[12pt,twoside]{article}
\usepackage{latexsym,amsmath,amssymb,hyperref}
\usepackage{graphicx}
\usepackage[tight]{subfigure}
\def\eqvsp{}  \newdimen\paravsp  \paravsp=1.3ex
\def\,{\mskip 3mu} \def\>{\mskip 4mu plus 2mu minus 4mu} \def\;{\mskip 5mu plus 5mu} \def\!{\mskip-3mu}
\def\dispmuskip{\thinmuskip= 3mu plus 0mu minus 2mu \medmuskip=  4mu plus 2mu minus 2mu \thickmuskip=5mu plus 5mu minus 2mu}
\def\textmuskip{\thinmuskip= 0mu                    \medmuskip=  1mu plus 1mu minus 1mu \thickmuskip=2mu plus 3mu minus 1mu}
\textmuskip
\def\beq{\eqvsp\dispmuskip\begin{equation}}    \def\eeq{\eqvsp\end{equation}\textmuskip}
\def\beqn{\eqvsp\dispmuskip\begin{displaymath}}\def\eeqn{\eqvsp\end{displaymath}\textmuskip}
\def\bqa{\eqvsp\dispmuskip\begin{eqnarray}}    \def\eqa{\eqvsp\end{eqnarray}\textmuskip}
\def\bqan{\eqvsp\dispmuskip\begin{eqnarray*}}  \def\eqan{\eqvsp\end{eqnarray*}\textmuskip}
\newenvironment{keywords}{\centerline{\bf\small
Keywords}\begin{quote}\small}{\par\end{quote}\vskip 1ex}
\def\paradot#1{\vspace{\paravsp plus 0.5\paravsp minus 0.5\paravsp}\noindent{\bf\boldmath{#1.}}}
\def\paranodot#1{\vspace{\paravsp plus 0.5\paravsp minus 0.5\paravsp}\noindent{\bf\boldmath{#1}}}
\def\req#1{(\ref{#1})}
\def\eps{\varepsilon}
\def\Euro{$\,$C$\!\!\!\!\!$\raisebox{0.2ex}{=}}
\def\SetR{I\!\!R}
\def\SetN{I\!\!N}
\def\SetZ{Z\!\!\!Z}
\def\qmbox#1{{\quad\mbox{#1}\quad}}
\def\E{{\bf E}}                         
\def\P{\text{Pr}}                       
\def\v{\boldsymbol}
\def\trp{{\!\top\!}}
\def\t{\theta}
\def\Loss{\mbox{Loss}}
\def\tr{\mbox{tr}}
\def\X{\v\t}

\begin{document}

\title{\vspace{-4ex}
\vskip 2mm\bf\Large\hrule height5pt \vskip 4mm
Featureless 2D-3D Pose Estimation by \\ Minimising an Illumination-Invariant Loss
\vskip 4mm \hrule height2pt}
\author{{\bf Srimal Jayawardena} and {\bf Marcus Hutter} and {\bf Nathan Brewer}\\[3mm]
\normalsize SoCS, RSISE, IAS, CECS \\[-0.5ex]
\normalsize Australian National University \\[-0.5ex]
\normalsize Canberra, ACT, 0200, Australia \\
\normalsize \texttt{\{srimal.jayawardena,marcus.hutter,nathan.brewer\}@anu.edu.au}
}
\date{3 November 2010}
\maketitle

\begin{abstract}
The problem of identifying the 3D pose of a known object from a
given 2D image has important applications in Computer Vision ranging
from robotic vision to image analysis.
Our proposed method of registering a 3D model of a known object on
a given 2D photo of the object has numerous advantages over existing
methods:
It does neither require prior training nor learning, nor knowledge of
the camera parameters, nor explicit point correspondences or
matching features between image and model.
Unlike techniques that estimate a partial 3D pose (as in an overhead
view of traffic or machine parts on a conveyor belt), our method
estimates the complete 3D pose of the object, and works on a single
static image from a given view, and under varying and unknown
lighting conditions.
For this purpose we derive a novel illumination-invariant distance
measure between 2D photo and projected 3D model, which is then
minimised to find the best pose parameters. Results for vehicle pose
detection are presented.
\def\contentsname{\centering\normalsize Contents}
{\parskip=-2.7ex\tableofcontents}
\end{abstract}

\vspace*{-2ex}
\begin{keywords}
illumination-invariant loss;
2D-3D pose estimation;
pixel-based; featureless;
optimisation.
{\let\thefootnote\relax\footnotetext{\noindent This work was supported by Control\Euro xpert.}}
\end{keywords}

\maketitle

\newpage
\section{Introduction}\label{secIntro}

Pose estimation is a fundamental component  of many computer
vision applications, ranging from robotic vision to intelligent
image analysis. In general, pose estimation refers to the
process of obtaining the location and orientation of an object.
However, the accuracy and nature of the pose estimate required
varies from application to application. Certain applications
require the estimation of the full 3D pose of an object, while
other applications require only a subset of the pose
parameters.

\paradot{Motivation}
The 2D-3D registration problem in particular is concerned with
estimating the pose parameters that describe a 3D object model
within a given 2D scene. An image/photograph of a known object
can be analysed in greater detail if a 3D model of the object
can be registered over it, to be used as a ground truth. As an
example, consider the case of automatically analysing a damaged
car  using a photograph. The focus of this work is to develop a
method to estimate the pose of a known 3D object model in a
given 2D image, with an emphasis on estimating the pose of
cars. We have the following objectives in mind.

\begin{itemize}
\item Use only a single, static image limited to a single view
\item Work with any unknown camera (without prior camera calibration)
\item Avoid user interaction
\item Avoid prior training / learning
\item Work under varying and unknown lighting conditions
\item Estimate the full 3D pose of the object (not a partial
    pose as in an overhead view of traffic or machine parts
    along a conveyor belt)
\end{itemize}

A 3D pose estimation method with these properties would also be
useful in remote sensing, automated scene recognition and
computer graphics, as it allows for additional information to
be extracted without the need for human involvement.

Many methods, including point correspondence based methods,
implicit shape model based methods and image gradient based
methods, have been developed to solve the pose estimation
problem. However, the methods identified in the literature do
not satisfy the objectives mentioned above, hence the necessity
of our novel method. A more detailed review of existing pose
estimation methods ranging over the past 30 years is presented
in Section~\ref{secRelatedWork}.

\paradot{Main contribution}
This paper presents a method which registers a known 3D model
onto a given 2D photo containing the modelled object while
satisfying the objectives outlined above. It does this by
measuring the closeness of the projected 3D model to the 2D
photo on a pixel (rather than feature) basis. Background and
unknown lighting conditions of the photo are major
complications, which prevent using a naive image difference
like the absolute or square loss as a measure of fit.

A major contribution of this paper is the novel ``distance''
measure in Section~\ref{secMatching3Dto2D} that does neither
depend on the lighting of the real scene in the photo nor on
choosing an appropriate lighting in the rendering of the 3D
model, hence does not require knowledge of the lighting.
Technically, we derive in Section~\ref{secLossFunction} a loss
function for vector-valued pixel attributes (of different
modality) that is invariant under linear transformations of the
attributes.

To analyse the nature of the developed loss function, we have
applied it to a series of test cases of varying complexity, as
detailed in Section \ref{secLossLandscape}. These test cases
indicate that for our target application the loss function is
well behaved and can be optimised using a standard optimisation
method to find an accurate pose match. As presented in Section
\ref{secOptim}, we achieve good pose recovery results in both
artificial and real world test cases using this optimisation
scheme.
In these optimisation tests, negative influence of the
background is attenuated by clipping the photo to the
projection of the 3D model when calculating the loss. Technical
aspects of the optimisation and loss calculation methods are
discussed in Section \ref{secTech}.

\section{Related Work}\label{secRelatedWork}

Model based object recognition has received considerable
attention in computer vision circles. A survey  by Chin and
Dyer \cite{modelbasedsurvey1986} shows that model based object
recognition algorithms generally fall into 3 categories, based
on the  type of object representation used - namely 2D
representations, 2.5D representations or 3D representations.

\paranodot{2D}
representations store the information of a particular 2D view
of an object (a characteristic view) as a model and use this
information to identify the object from a 2D image. Global
feature methods have been used by Gleason and Algin
\cite{gleason1979} to identify objects like spanners and nuts
on a conveyor belt. Such methods use features such as the area,
perimeter, number of holes visible and other global features to
model the object. Structural features like boundary segments
have been used by Perkins \cite{perkins1978} to detect machine
parts using 2D models. A relational graph method has been used
by Yachida and Tsuji \cite{yachida1977} to match objects to a
2D model using graph matching techniques. These 2D
representation-based algorithms require prior training of the
system using a `show by example' method.

\paranodot{2.5D}
approaches are also viewer centred, where the object is known
to occur in a particular view. They differ from the 2D approach
as the model stores additional information such as intrinsic
image parameters and surface-orientation maps. The work done by
Poje and Delp \cite{poje1982} explain the use of intrinsic
scene parameters in the form of range (depth) maps and needle
(local surface orientation) maps. Shape from shading
\cite{horn1975} and photometric stereo
\cite{woodham1978photometric} are some other examples of the
use of the 2.5D approach used for the recognition of industrial
parts. A range of techniques for such 2D/2.5D representations
are described by Forsythe and Ponce \cite{forsythe}, by posing
the object recognition problem as a correspondence problem.
These methods obtain a hypothesis based on the correspondences
of a few matching points in the image and the model. The
hypothesis is validated against the remaining known points.

\paranodot{3D}
approaches are utilised in situations where the object of
interest can appear in a scene from multiple viewing angles. Common
3D representation approaches can be either an `exact representation'
or a `multi-view feature representation'. The latter method uses a
composite model consisting of 2D/2.5D models for a limited set of
views. Multi-view feature representation is used along with the
concept of generalised cylinders by Brooks and Binford
\cite{brookes1981} to detect different types of industrial motors in
the so called ACRONYM system. The models used in the exact
representation method, on the contrary, contain an exact
representation of the complete 3D object. Hence a 2D projection of
the object can be created for any desired view. Unfortunately, this
method is often considered too costly in terms of processing time.

\paradot{Limitations}
The 2D and 2.5D representations are insufficient for general
purpose applications. For example, a vehicle may be
photographed from an arbitrary view in order to indicate the
damaged parts. Similarly, the 3D multi-view feature
representation is also not suitable, as we are not able to
limit the pose of the vehicle to a small finite set of views.
Therefore, pose identification has to be done using an exact 3D
model. Little work has been done to date on identifying the
pose of an exact 3D model from a single 2D image. Huttenlocher
and Ullman \cite{huttenlocher1990recognizing}  use a 3D model
that contains the locations of edges. The edges/contours
identified in the 2D image are matched against the edges in the
3D model to calculate the pose of the object. The method has
been implemented for simple 3D objects. However, this method
will not work well on objects with rounded surfaces without
clearly identifiable edges.

\paradot{Implicit Shape Models}
Recent work  by Arie-Nachimson and Ronen Basri
\cite{implicitshapepose} makes use of `implicit shape models'
to recognise 3D objects from 2D images. The model consists of a
set of learned features, their 3D locations and the views in
which they are visible. The learning process is further refined
using factorisation methods. The pose estimation consists of
evaluating the transformations of the features that give the
best match. A typical model requires around 65 images to be
trained. There are many different types of cars in use and new
car models are manufactured quite frequently. Therefore, any
methodology that requires training car models would be
laborious and time consuming. Hence, a system that does not
require such training is preferred for the problem at hand.

\paradot{Image gradients}
Gray scale image gradients have been used to estimate the 3D
pose in traffic  video footage from a stationary camera by
Kollnig and Nagel \cite{3dposegrayval}. The method compares
image gradients instead of simple edge segments, for better
performance. Image gradients from projected polyhedral models
are compared against image gradients in video images. The pose
is formulated using 3 degrees of freedom; 2 for position and 1
for angular orientation. Tan and Baker \cite{tan2000efficient}
use image gradients and a Hough transform based algorithm for
estimating  vehicle pose in traffic scenes, once more
describing the pose via 3 degrees of freedom. Pose estimation
using 3 degrees of freedom  is adequate for traffic image
sequences, where the camera position remains fixed with respect
to the ground plane. This approach does not provide a full pose
estimate required for a general purpose application.

\paradot{Feature-based methods}
Work done by \cite{david2004softposit} and later by
\cite{moreno2008pose} attempt to simultaneously solve  the pose
and point correspondence problems. The success of these methods
are affected by the quality of the features extracted from the
object, which is non-trivial with objects like cars. Our method
on the contrary, does not depend on feature extraction.

\paranodot{Distance metrics}
can be used to represent a distance between two data sets, and
hence give a measure of their similarity. Therefore, distance
metrics can be used to measure similarity between different 2D
images, as well as 2D images and projections of a 3D model. A
basic distance metric would be the \emph{Euclidian Distance} or
the 2-norm $||\cdot||_2$. However, this has the disadvantage of
being dependant on the scale of measurement. The
\emph{Mahalanobis Distance} on the other hand, is a
scale-invariant distance measure. It is defined as
\beqn
  ||x-y||_{C^{-1}} \equiv \sqrt{(x-y)^\trp C^{-1}(x-y)}
\eeqn
for random vectors $x$ and $y$ with a covariance matrix of $C$. The
Mahalanobis distance will reduce to the Euclidean distance when the
covariance matrix is the identity matrix ($C=I$). The Mahalanobis
distance is used by Xing et al.\ \cite{xing2003distance} for
clustering. It is also used by Deriche and Faugeras
\cite{deriche1990tracking} to match line segments in a sequence of
time varying images.

\section{Matching 3D Models with 2D Photos}\label{secMatching3Dto2D}

We describe our approach of matching 3D models to 2D photos in this
section using a novel illumination-invariant loss function. A
detailed derivation of the loss is provided in
Section~\ref{secLossFunction}.

\paradot{The problem}
Assume we want to match a 3D model ($M$) to a 2D photo ($F$) or
vice versa. More precisely, we have a 3D model (e.g.\ as a
triangulated textured surface) and we want to find a projection
$\theta$ for which the rendered 2D image $M_\theta$ has the
same perspective as the 2D photo $F$. As long as we do not know
the lighting conditions of $F$, we cannot expect $F$ to be
close to $M_\theta$, even for the correct $\theta$. Indeed, if
the light in $F$ came from the right, but the light shines on
$M$ from the left, $M_\theta$ may be close to the negative of
$F$.

\paradot{Setup}
Formally, let $P=\SetZ_{n_x}\times
\SetZ_{n_y}=\{1,...,n_x\}\times\{1,...,n_y\}$ be the set of
$|P|$ (integer) pixel coordinates, and $p=(x,y)\in P$ be a
pixel coordinate. Alternatively a smaller region of interest
may be used for $P$ instead of $\SetZ_{n_x}\times \SetZ_{n_y}$
as explained in Section~\ref{ClipBG}.
Let $F:P\to\SetR^n$ be a photo with $n$ real pixel attributes,
and $M_\theta:P\to\SetR^m$ a projection of a 3D object to a 2D
image with $m$ real pixel attributes. The attributes may be
colours, local texture features, surface normals, or else. In
the following we consider the case of grey-level photos
($n=1$), and for reasons that will become clear, use surface
normals and brightness $(m=4)$ of the (projected) 3D model.

\paradot{Lambertian reflection model}
A simple Lambertian reflection model is not realistic enough to
result in a zero loss on real photos, even at the correct pose.
Nevertheless (we believe and experimentally confirm that) it
results in a minimum at the correct pose, which is sufficient
for matching purposes. We use Phong shading without specular
reflection for this purpose \cite{Foley:95}. Let
$I_{a/d}\in\SetR$ be the global ambient/diffuse light
intensities of the 3D scene, and $\v L\in\SetR^3$ be the
(global) unit vector in the direction of the light source (or
their weighted sum in case of multiple sources). For reasons to
become clear later, we introduce an extra illumination offset
$I_0\in\SetR$ (which is 0 in the Phong model). For each surface
point $p$, let $k_{a/d}(p)\in\SetR$ be the ambient/diffuse
reflection constants (intrinsic surface brightness) and
$\v\phi(p)\in\SetR^3$ be the unit (interpolated) surface
``normal'' vector. Then the apparent intensity $I$ of the
corresponding point $p$ in the projection $M_\theta(p)$ is
\cite{Foley:95}
\beqn
  I(p) = k_a(p) I_a \!+\! k_d(p)(\v L^\trp\v\phi(p))I_d \!+\! I_0
    \equiv A\!\cdot\!M_\theta(p) \!+\! b
\eeqn
The last expression is the same as the first, just written in a
more covariant form:
$M_\theta(p):=(k_a(p),k_d(p)\v\phi(p))^\trp\in\SetR^4$ are the
known surface (dependent) parameters, and $A:=(I_a,I_d\v
L^\trp)\in\SetR^{4\times 1}$ are the four (unknown) global
illumination constants, and $b=I_0$. Since $I(\cdot)$ is linear
in $A$ and $M_\theta(\cdot)$, any rendering is a simple global
linear function of $M_\theta(p)$.
This model remains exact even for multiple light sources
and can easily be generalised to color models and color photos.

\paradot{Illumination invariant loss}
We measure the closeness of the projected 3D model $M_\t$ to
the 2D photo $F$ by some distance $D(F,A M_\t+b)$, e.g.\ square
or absolute or Mahalanobis. We do not want to assume any extra
knowledge like the lighting conditions $A$ under which the
photo has been taken, which rules out a direct use of $D$.
Ideally we want a ``distance'' between $F$ and $M$ that is
independent of $A$ and is zero if and only if there exists a
lighting condition $A$ such that $F$ and $AM_\t+b$ coincide.

Indeed, this is possible, if (rather than defining $M_\theta$ as some
$A$-dependent rendered projection of $M$) we use $A$-independent
brightness and normals $M_\t$ as pixel features as defined above,
and define a linearly invariant distance as follows: Let
\beqn
  \bar F    \;:=\; {1\over |P|}\sum_{p\in P} F(p)\;\in\SetR \qmbox{and}
  \bar M_\t \;:=\; {1\over |P|}\sum_{p\in P} M_\t(p)\;\in\SetR^4
\eeqn
be the average attribute values of photo and projection, and
\beqn
  C_{F M_\t} := {1\over |P|}\sum_{p\in P}(F(p)-\bar F)(M_\t(p)-\bar M_\t)^\trp\;\in\SetR^{1\times 4}
\eeqn
be the cross-covariance matrix between $F$ and $M_\t$ and
similarly $C_{M_\t F}=C_{FM_\t }^\trp\in\SetR^{4\times 1}$ and
the covariance matrices $C_{FF}\in\SetR^{1\times 1}$ and
$C_{M_\t M_\t}\in\SetR^{4\times 4}$. With this notation we can
define the following distance or loss function between $F$ and
$M_\t$:
\beq\label{defLIL}
  \Loss(\theta) := \min\{n,m\}
  - \tr[C_{FM_\theta}C_{M_\theta M_\theta}^{-1}C_{M_\theta F}C_{FF}^{-1}]
\eeq
Obviously this expression is independent of $A$. In the next
section we show that it is invariant under regular linear
transformation of the pixel/attribute values of $F$ and
$M_\theta$ and zero if and only if there is a perfect linear
transformation of the pixel values from $M_\theta$ to $F$. This
makes it unnecessary to know the exact surface reflection
constants of the object ($k_{a/d}(p)\in\SetR$).
We will actually derive
\beqn
  \Loss(\theta) = \smash{\min_{A,b}}\,D_{\text{Mahalanobis}}(F,A\!\cdot\!M_\t+b)
\eeqn
This implies that $\Loss(\theta)$ is zero if and only if there
is a lighting $A$ under which $F$ and $M_\t$ coincide, which we
desired.

\section{Derivation of Invariant Loss Function}\label{secLossFunction}

A detailed derivation of the loss function is given in this section.
Although together with Section \ref{secMatching3Dto2D}, this is a
main novel contribution of this paper, it may be skipped over by
the more application-oriented reader without affecting the
continuity of the rest of the paper.

\paradot{Notation}
Using the notation of the previous section, we measure the
similarity of photo $F:P\to\SetR^n$ and projected 3D model
$M_\theta:P\to\SetR^m$ (returning to general $n,m\in\SetN$) by some
loss:
\beq\label{defLoss}
  \Loss(\theta) := D(F,M_\theta)
  := {1\over|P|}\smash{\sum_{p\in P}} d(F(p),M_\theta(p))
\eeq
where $d$ is a distance measure between corresponding pixels of the
two images to be determined below. A very simple, but as discussed
in Section \ref{secRelatedWork} for our purpose unsuitable,
choice in case of $m=n$ would be the square loss
$d(F(p),M_\theta(p))=||F(p)-M_\theta(p)||_2^2$.

It is convenient to introduce the following probability
notation: Let $\omega$ be uniformly distributed\footnote{With a
non-uniform distribution one can easily weigh different pixels
differently.} in $P$, i.e.\ $\P[\omega]=|P|^{-1}$. Define the
vector random variables $X:=F(\omega)\in\SetR^n$ and
$Y:=M_\theta(\omega)\in\SetR^m$. The expectation of a function
of $X$ and $Y$ then is
\beqn
  \E[g(X,Y)]:={1\over|P|}\sum_{\omega\in P}g(X(\omega),Y(\omega))
\eeqn
With this notation, \req{defLoss} can be written as
\beqn
  \Loss(\theta) = D(X,Y) = \E[d(X,Y)]
\eeqn

\paradot{Noisy (un)known relation}
Let us now assume that there is some (noisy) relation $f$ between
(the pixels of) $F$ and $M_\theta$, i.e.\ between $X$ and $Y$:
\beqn
  Y=f(X)+\eps,\qquad \eps=\mbox{noise}
\eeqn
If $f$ is known and $\eps$ is Gaussian, then
\beqn
  D_f(X,Y)=\E[||f(X)-Y||_2^2]
\eeqn
is an appropriate distance measure for many purposes. In case $F$
and $M_\theta$ are from the same source (same pixel attributes, lighting
conditions, etc) then $f=$identity is appropriate and we get the
standard square loss. In many practical applications, $f$ is not the
identity and furthermore unknown (e.g.\ mapping gray models to real
color photos of unknown lighting condition). Let us assume $f$
belongs to some set of functions $\cal F$. $\cal F$ could be the set
of all functions or just contain the identity or anything in between
these two extremes. Then the ``true/best'' $f$ may be estimated by
minimising $D_f$ and substituting into $D_f$:
\beqn
  f_{best}=\arg\min_{f\in\cal F} D_f(X,Y),\qquad
  D(X,Y):= \min_{f\in\cal F}D_f(X,Y)
\eeqn
Given $\cal F$, $D$ can in principle be computed and measures
the similarity between $X$ and $Y$ for unknown $f$.
Furthermore, $D$ is invariant under any transformation $X\to
g(X)$ for which ${\cal F}\circ g=\cal F$.

\paradot{Linear relation}
In the following we will consider the set of linear relations
\beqn
  {\cal F}_{lin} \;:=\; \{f:f(X)=AX+b,A\in\SetR^{m\times n},b\in\SetR^m\}
\eeqn
For instance, a linear model is appropriate for mapping color
to gray images (same lighting), or positives to negatives. For
linear $f$, $D$ becomes
\beqn
  D(X,Y) \;=\; \min_{A\in\SetR^{m\times n}}\min_{b\in\SetR^m}
               \E[||AX+b-Y||_2^2]
\eeqn
Good news is that this distance is invariant under all regular
linear reparametrisations of $X$, i.e.\ $D(X,Y)=D(AX+b,Y)$ for
all $b$ and all non-singular $A$. Unfortunately, $D$ is not
symmetric in $X$ and $Y$ and in particular not invariant under
linear transformations in $Y$. Assume the components
$(Y_1,...,Y_m)^\trp$ are of very different nature ($Y_1$=color,
$Y_2$=angle, $Y_3$=texture), then the 2-norm $||Y||_2^2=Y^\trp
Y=Y_1^2+...+Y_m^2$ compares apples with pears and makes no
sense. A standard solution is to normalise by variance, i.e.\
use $\sum_i Y_i^2/\sigma_i^2$, where
$\sigma_i^2=\E[Y_i^2]-\E[Y_i]^2$, but this norm is (only)
invariant under component scaling.

\paradot{Linearly invariant distance}
To get invariance under general linear transformations, we have to
``divide'' by the covariance matrix
\beqn
  C_{YY} \;:=\; \E[(Y-\bar Y)(Y-\bar Y)^\trp],\quad \bar Y:=\E[Y]
\eeqn
The Mahalanobis norm (cf.\ Section \ref{secRelatedWork})
\beqn
  ||Y||_{C_{YY}^{-1}}^2 \;:=\; Y^\trp C_{YY}^{-1} Y
\eeqn
is invariant under linear {\em homogenous} transformations, as can be seen from
\beqn
  ||AY||_{C_{AY,AY}^{-1}}^2
  \;\equiv\; Y^\trp A^\trp C_{AY,AY}^{-1}AY
  \;=\; Y^\trp C_{YY}^{-1} Y
  \;\equiv\; ||Y||_{C_{YY}^{-1}}^2
\eeqn
where we have used $C_{AY,AY}=A C_{YY} A^\trp$.

The following distance is hence invariant under {\em any}
non-singular linear transformation of $X$ and any non-singular
(incl.\ non-homogenous) linear transformation of $Y$:
\beq\label{defDinv}
  D(X,Y) \;:=\; \min_{A\in\SetR^{m\times n}}\min_{b\in\SetR^m}
               \E[||AX+b-Y||_{C_{YY}^{-1}}^2]
\eeq

\paradot{Explicit expression}
Since the norm$^2$ is quadratic in $A$ and $b$, the minimisation can
be performed explicitly, yielding
\beq\label{Abmin}
  b=b_{min}:=\bar Y-A_{min}\bar X \qmbox{and}
  A=A_{min}:=C_{YX}C_{XX}^{-1},\qmbox{where}
\eeq
\beqn
  C_{XY} \;:=\; \mbox{Cov}(X,Y)
  \;=\; \E[(X-\bar X)(Y-\bar Y)^\trp], \quad
  \bar X:=\E[X]
\eeqn
and similarly $C_{YX}=C_{XY}^\trp$ and $C_{XX}$. Inserting
\req{Abmin} back into \req{defDinv} and rearranging terms gives
\beqn
  D(X,Y) \;=\; \tr[1\!\!1-C_{YX}C_{XX}^{-1}C_{XY}C_{YY}^{-1}]
  \;=\;   m - \tr[C_{XY}C_{YY}^{-1}C_{YX}C_{XX}^{-1}].
\eeqn
This explicit expression shows that $D$ is also nearly
symmetric in $X$ and $Y$. The trace is symmetric but $m$ is
not. For comparisons, e.g.\ for minimising $D$ w.r.t.\
$\theta$, the constant $m$ does not matter. Since the trace can
assume all and only values in the interval $[0,\min\{n,m\}]$,
it is natural to symmetrize $D$ by
\beqn
  \min\{D(X,Y),D(Y,X)\} \;=\;
  \min\{n,m\} - \tr[C_{XY}C_{YY}^{-1}C_{YX}C_{XX}^{-1}]
\eeqn
Returning to original notation, this expression coincides with
the loss \req{defLIL}. It is hard to visualize this loss, even
for $n=1$ and $m=4$, but the special case $m=n=1$ is
instructive, for which the expression reduces to
\beqn
  D(X,Y)=1-\mbox{corr}^2(X,Y), \qmbox{where}
  \mbox{corr}(X,Y) = {\mbox{Cov}(X,Y)\over \sigma_X\sigma_Y}
\eeqn
is the correlation between $X$ and $Y$. The larger the
(positive or {\em negative}) correlation, the more similar the
images and the smaller the loss. For instance, a photo has
maximal correlation with its negative.

\section{Practical Behaviour of the Loss Function}\label{secLossLandscape}

In this section, we explore the  nature of the loss function derived
in Section~\ref{secLossFunction} for real and artificial
photographs, together with a pose representation specific to
vehicles.

\begin{figure}[t!]
\centering
\subfigure[Pose representation $\X$]{\label{fig:poseRepresentationX}
\includegraphics[width=0.48\textwidth]{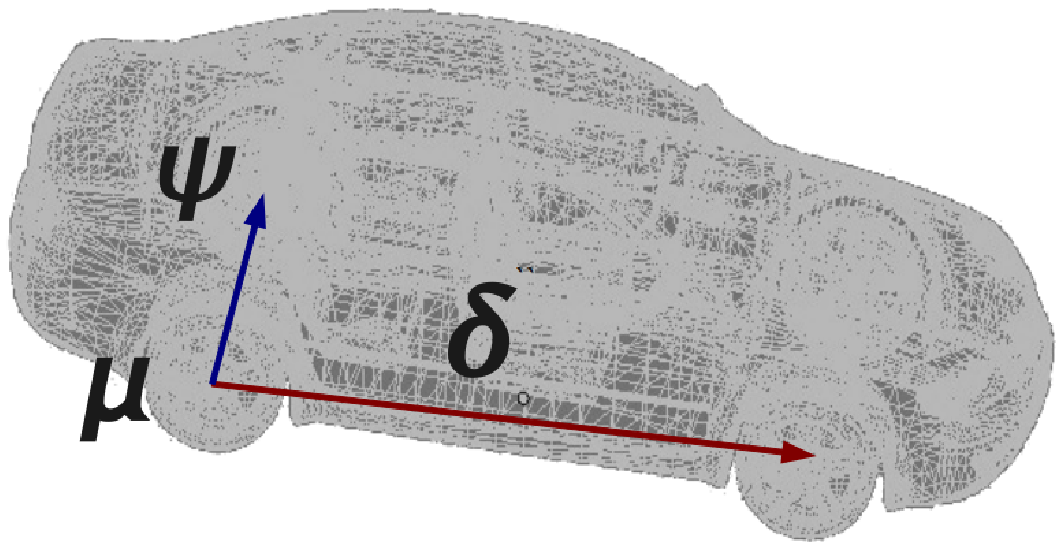} }
\subfigure[Pose deviations]{\label{fig:startingPoseDeviations50}
    \includegraphics[width=0.48\textwidth,height=0.3\textwidth]{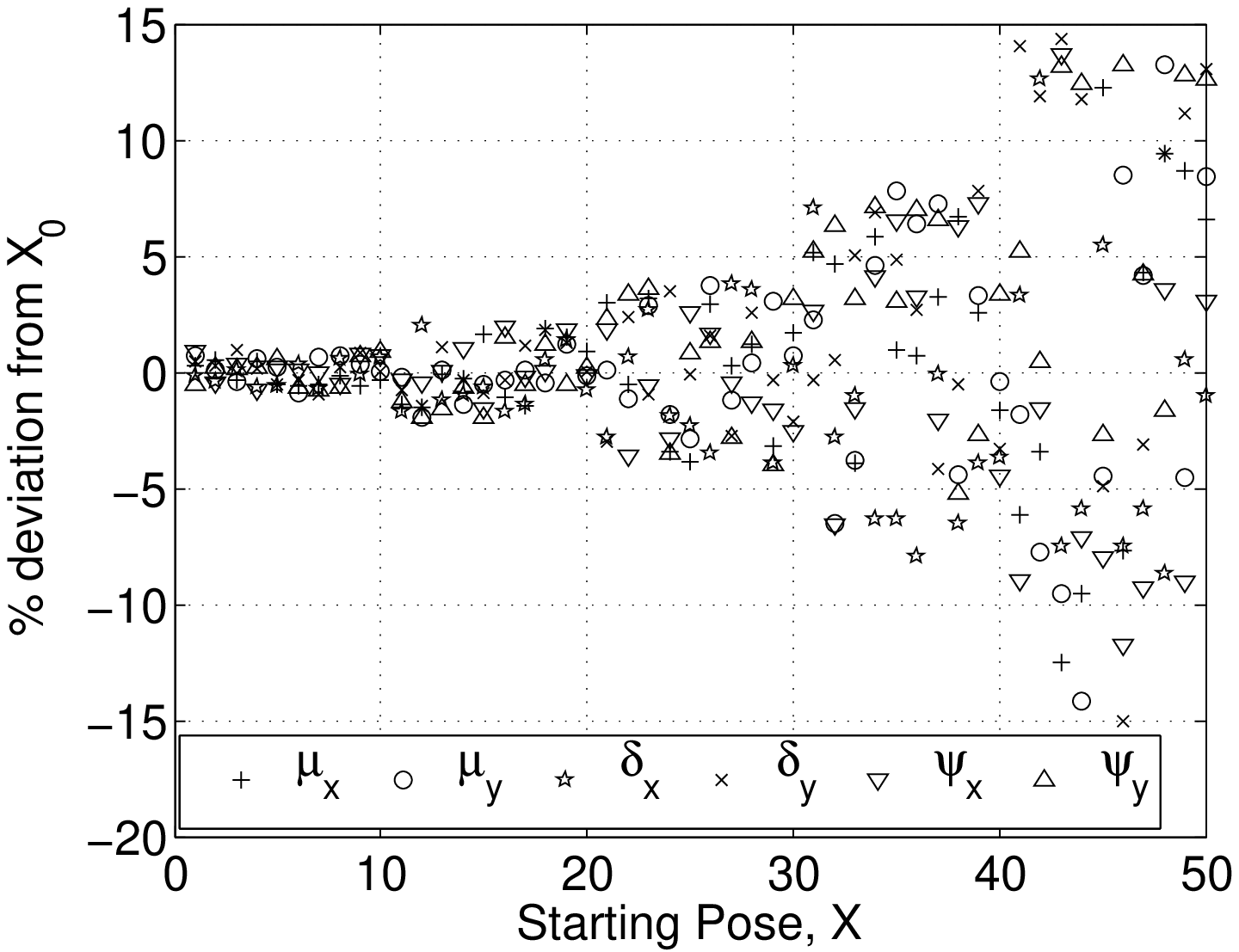}}

\caption[Pose representation and initial pose deviations] {Figure
\ref{fig:poseRepresentationX} shows the pose representation $\X$
used for 3D car models. We use the rear wheel center $\v\mu$,
the vector between the wheel centres $\v\delta$ and unit vector $\v\psi$ in
the direction of the rear wheel axle. Figure
\ref{fig:startingPoseDeviations50} shows the deviations of the
randomly generated starting poses used for reliability testing. The
test cases were generated to fall within percentage deviations in
the ranges of 1\%, 2\%, 4\%, 8\% and 16\% from the true pose
$\X_0$ in the image. Poses 1-10 have a deviations of 1\%, poses 11-20 have a
deviation of 2\% etc. }
\end{figure}

\paradot{Representation of the pose}
It is important to select a pose representation that suitably
describes the 3D model that is being matched. Careful selection
of pose parameters can enhance the ability of the optimisation
to find the best match, and can allow object detection or
coarse alignment methods, such as that presented in
\cite{hutter2009matching} to specify a starting pose for the
optimisation. We use the following pose representation for 3D
car models, temporarily neglecting the effects of perspective
projection:
\beq\label{eqn:PoseX}
  \X := \left( \mu_x, \mu_y, \delta_x, \delta_y, \psi_x, \psi_y \right)
\eeq
$\v\mu=(\mu_x, \mu_y)$ is the visible rear wheel center of the
car in the 2D projection. $\v\delta=(\delta_x, \delta_y)$ is
the vector between corresponding rear and front wheel centres
of the car in the 2D projection.
$\v\psi = \left( \psi_x, \psi_y, \psi_z \right)$ is a unit
vector in the direction of the rear wheel axle of the 3D car
model. Therefore, $\psi_z = -\sqrt{1 - \smash{\psi_x^2 -
\psi_y^2}}$ and need not be explicitly included in the pose
representation $\X$. This representation is illustrated in
Figure~\ref{fig:poseRepresentationX}.

\begin{figure}[t!]
\centering\hspace{-3ex}
\subfigure[Artificial photo]{\label{fig:F1Surf}
     \includegraphics[width=0.5\textwidth]{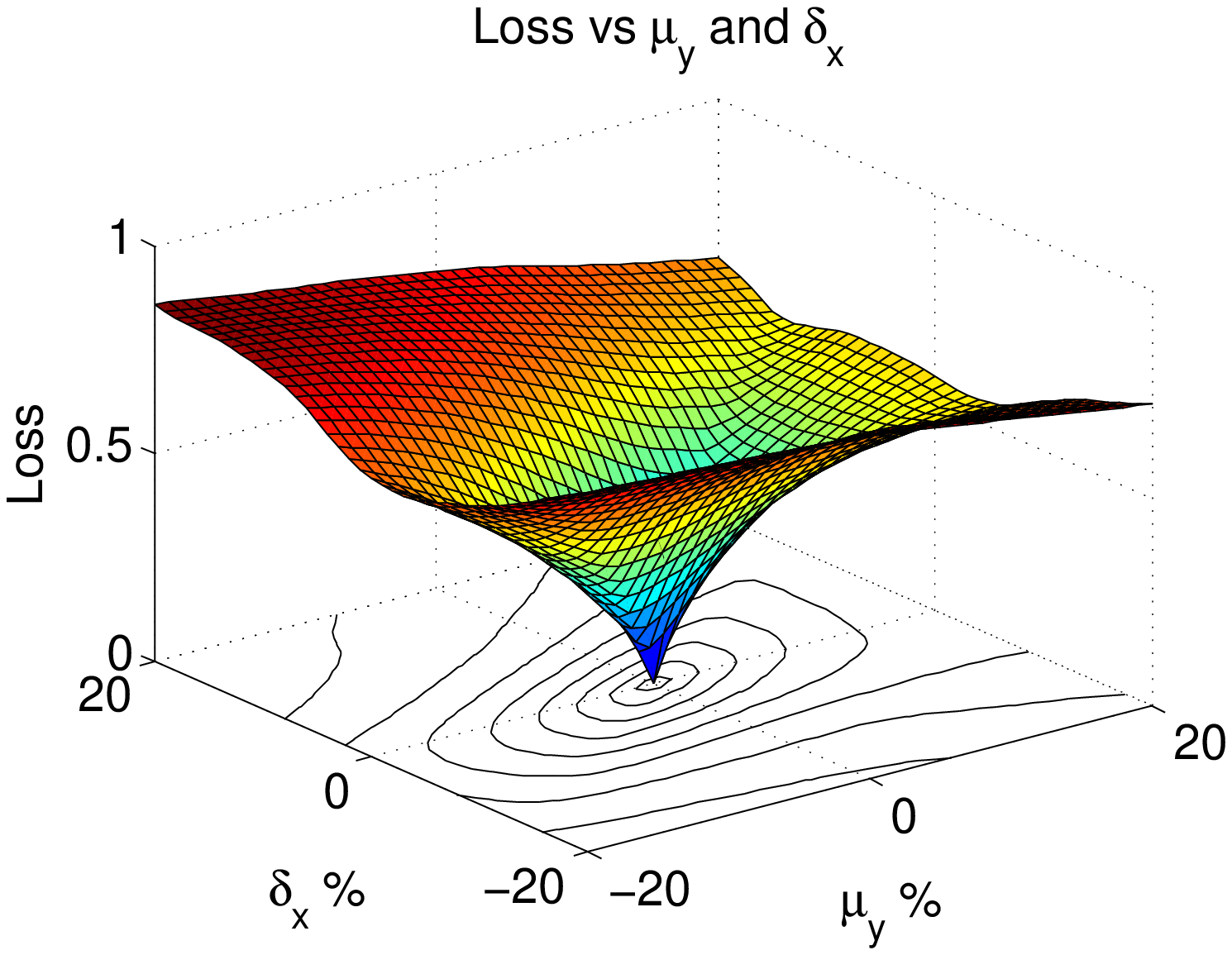}}
\hspace{-2ex}
\subfigure[Real photo]{\label{fig:LosslandscapeRealPhoto}
    \includegraphics[width=0.5\textwidth]{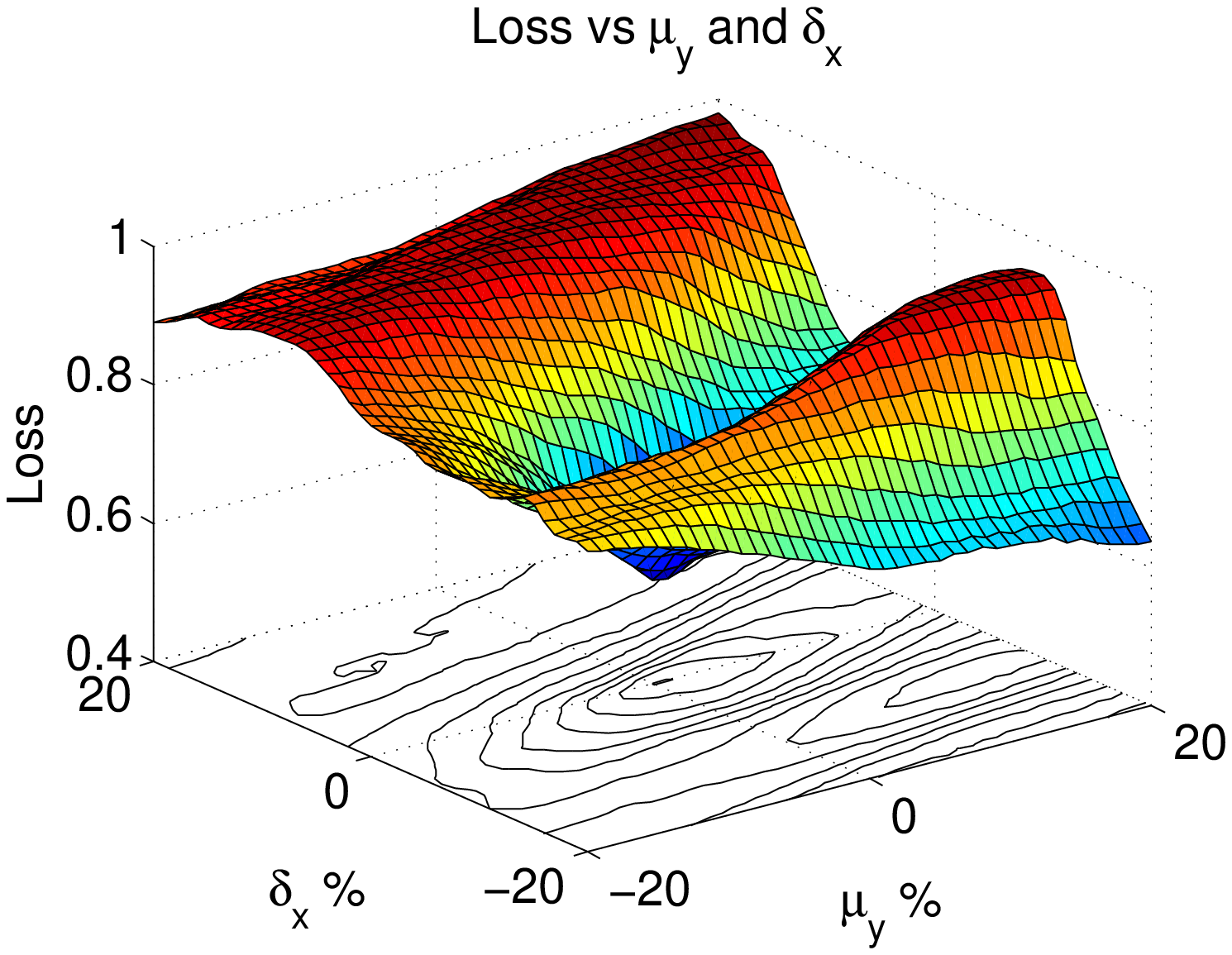}}
\hspace{-1ex}
\caption[Loss landscapes for artificial and real photos]{ {\bf Loss
landscapes for artificial and real photos.} The six dimensional loss
function was visualised by plotting its variation with a pair of
pose parameters at a time. Based on our pose representation this
results in fifteen plots. The variation of the loss function with a
pair of pose parameters are shown for an artificial photograph and
a real photograph. The nature of the loss function for real
photographs makes it more difficult to find the global minimum
(hence the correct pose) than for artificial photographs.}
\label{fig:LossLandscapeArtificialAndRealPhotos}
\end{figure}

\paradot{Artificial photographs}
To understand the behaviour of the loss function, we have
generated loss landscapes for artificial images of 3D  models.
To produce these landscapes, an artificial photograph was
generated by projecting the 3D model at a known pose $\X_0$
with Phong shading. We then vary the pose parameters, two at a
time about $\X_0$ and find the value of the loss function
between this altered projection and the ``photograph'' taken at
$\X_0$. These loss values are recorded, allowing us to
visualise the behaviour of the loss function by observing
surface and contour plots of these values. The unaltered pose
values should project an image identical to the input
photograph, giving a loss of zero according to the loss
function derived in Section~\ref{secLossFunction}, with a
higher loss exhibited at other poses. The variation of the loss
with respect to a pair of pose parameters is shown in Figure
\ref{fig:F1Surf}. It can be seen from these loss landscapes
that the loss has a clear minimum at the initial pose $\X_0$.
The loss values increase as these pose parameters deviate away
from $\X_0$, up to $\pm 20\%$. From this data, we are able to
see that the minimum corresponding to $\X_0$ can be considered
a global minimum for all practical purposes. The shape of the
surface plots was similar for all other parameter pairs,
indicating that the full 6D landscape of the loss function
should similarly have a global minimum at the initial pose,
allowing us to find this point using standard optimisation
techniques, as demonstrated in Section~\ref{secOptim}.

\paradot{Loss landscape}
The landscape of the loss function was analysed for real
photographs by varying the pose parameters of the model about a
pose obtained by manually matching the 3D car model to a real
photograph. The variation was plotted by taking a pair of pose
parameters at a time over the entire set of pose parameters. A
loss landscape obtained by varying $\mu_y$ and $\delta_x$ for a
real photograph is shown in Figure
\ref{fig:LosslandscapeRealPhoto}. The variation of the loss
function for other pose parameter pairs were found to be
similar. Although a global minimum exists at the best pose of
the real photograph, the nature of the loss function surface
makes it more difficult to optimise when compared to artificial
photos (Figure \ref{fig:F1Surf}). In particular, one can
observe local minima in the periphery of the landscape, and the
full 6D landscape is considerably more complex.

\section{Optimising Vehicle Pose using the Loss Function}\label{secOptim}

As explained in Section~\ref{secLossFunction}, the correct pose
parameters $\v{\theta_{opt}}$ will  give  the lowest loss
value. The loss function landscape, as discussed in
Section~\ref{secLossLandscape}, shows that $\v{\theta_{opt}}$
corresponds to the global minimum of the loss function.
Therefore, an optimisation was performed on the loss function
to obtain $\v{\theta_{opt}}$ based on the pose representation
$\X$ (Equation \ref{eqn:PoseX}). The optimisation strategy  and
its reliability in different scenarios is discussed in this
section.

\paradot{The Optimiser}
To immunise the optimisation from pixel quantisation artefacts and
noise in the images, direct search methods that do not calculate the
derivative of the loss function were considered.
%
The optimisation was performed using the well known
\emph{Downhill Simplex Method (DS)}
\cite{nelder1965simplex,nr2,matlab}, owing to its efficiency
and robustness. When optimising an $n$-dimensional function
with the DS method, a so called \emph{simplex} consisting of
$n+1$ points is used to traverse the $n$-dimensional search
space and find the optimum.

The reliability of the optimisation is adversely affected by the
existence of local minima. Fortunately, the \emph{Downhill Simplex}
method has a useful property. In most cases, if the simplex is
reinitialised at the pose parameters of the local minimum and the
optimisation is performed again, the solution converges to the
global minimum.
%
Proper parameterisation is important for the optimiser to give
good results. We have used a normalised pose parameterisation
as follows.

\paradot{Normalised pose parameters}
Normalisation gives each pose parameter a comparable range
during optimisation. The normalised pose $\v{\X_N}$ was
obtained by normalising $\v\mu$ and $\v\delta$ w.r.t.\ the
dimensions of the photograph.
\beqn
  \v{\X_N} = \left( \frac{\mu_x}{I_W} , \frac{\mu_y}{ I_H} , \frac{\delta_x}{ I_W}, \frac{\delta_y }{I_H}  , \psi_{x}, \psi_{y} \right)
\eeqn
$I_W , I_H $ are the width and height of the photograph (2D image).
$\v\psi$ is a unit vector and does not require normalisation.

\paradot{Initialisation}
The downhill simplex method, like all optimisation techniques,
requires a reasonable starting position. There are many methods for
selecting a starting point, from repeated random initialisation to
structured partitioning of the optimisation volume. A disadvantage
of these methods is that they require a number of optimisation runs
to locate the optimal point, which can take significant time.
Depending on the application, it may be possible to develop a coarse
location method which provides an estimate of the optimal pose.

The wheel match method described by Hutter and Brewer
\cite{hutter2009matching} is one such method, providing an initial
match for a vehicle pose if the vehicle's wheels are visible. Wheel
match estimations using this method generally locate the wheel
centres with a high degree of accuracy, but perform less effectively
when determining the axle direction. This indicates that it may be
possible to perform staged optimisation, attempting to fix some
parameters before others. In general, parameter estimation using
this method is within 5-15\% of the true value. This initial pose
selection is sufficient for our purposes.

\paradot{Reliability of the pose estimate}
Tests were carried out to asses the reliability of the pose
estimate. We first generated synthetic ``photographs'' by
rendering a 2D projection of a 3D car model at a known pose
$\v{\X_0}$. The optimisation was performed from initial poses
$\v\X$ that had a known deviation from $\v{\X_0}$. Test poses
were selected at 1\%, 2\%, 4\%, 8\% and 16\% deviation from the
known initial parameters so as to investigate a large
hyper-volume in 6D space. The parameter values for 50 such
random starting poses, 10 for each range, are shown in Figure
~\ref{fig:startingPoseDeviations50}. The reliability for each
percentage range was defined as the proportion of correct
matches in that deviation range. Exact pose recovery for
synthetic images and better than careful manual tuning for real
photos were regarded as {\em correct}. With this definition, a
reliability of 1 indicates that all test cases in the range
converged. A reliability of 0 indicates that none of the test
cases in the range converged.

\begin{figure}[t!]
\begin{minipage}{0.49\textwidth}
\includegraphics[width=\textwidth]{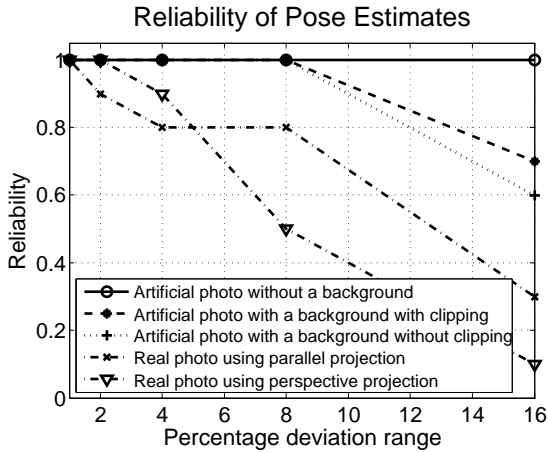}
\end{minipage}
\begin{minipage}{0.49\textwidth}\vspace*{-3ex}
\caption[Results of the reliability tests] {{\bf Results of the
reliability tests.} The graph shows the results of the reliability
tests based on initial poses with percentage deviations as shown in
Fig.\ref{fig:startingPoseDeviations50}. Results for an artificial
photograph (a projection of the 3D model), an artificial photograph
with a real background superimposed and a real photograph are
included.} \label{fig:combinedResultsGraph}
\end{minipage}
\end{figure}

\paradot{Reliability for artificial images}
To ensure that the selected optimisation method is appropriate,
we first investigate a simple case in which an artificial image
with known parameters is constructed and used to validate the
optimisation method. The reliability of the optimisation (with
simplex re-initialisations) was found to be 100\% (Figure
\ref{fig:combinedResultsGraph}) for initial poses with up to a
16\% deviation from the matching pose.

\paradot{Reliability for artificial images with real backgrounds}
Next we rendered artificial car models on a real background
photo, and performed the same reliability tests. Allowing
simplex re-initialisations preserved a 100\% convergence up to
the 8\% deviation range (Figure
\ref{fig:combinedResultsGraph}), although the simplex did not
converge for certain starting poses  at a 16\% deviation. This
shows that the effects of a real background can deteriorate the
reliability of the pose estimate for higher deviations. In
order to address this issue, a further refinement of the
algorithm was made by clipping the background in the photograph
when calculating the loss.

\paradot{Clipping the background}\label{ClipBG}
The methodology used to lower the effects of the image
background is as follows: Pixels in the projected image that do
not correspond to points of the 3D model were treated as
background. These pixels do not have surface normal components
as they do not belong to the 3D model. Therefore, they can
easily be filtered out by identifying pixels in the projected
image that have null values for all three components $(x,y,z)$
of the surface normal. Only the remaining pixels $P=\{p\in
\SetZ_{n_x} \times \SetZ_{n_y} : \v\phi(p)\neq \v0\}$ were
considered for the loss calculation (Figure
\ref{fig:combinedResultsGraph}).

\paradot{Reliability for real images using parallel projection}
The reliability of the pose
estimates on a real car photo are shown in Figure
\ref{fig:combinedResultsGraph}. Correct pose estimates with a
high reliability were obtained for starting poses up to
an 8\% deviation.

\paradot{Reliability for real images using perspective projection}
The distance from the camera to the projection plane in the
OpenGL perspective projection model was used as a seventh
parameter when optimising using perspective projection. The
extra pose parameter makes the optimisation harder at higher
deviations as seen in the reliability graph in Figure
\ref{fig:combinedResultsGraph}. The reliability of the pose
estimate may be further improved by using more sophisticated
optimisation methods.

\begin{figure}[t!]
\centering
\subfigure[Starting pose]{\label{fig:realphotoInitialPose}
     \includegraphics[width=0.48\textwidth]{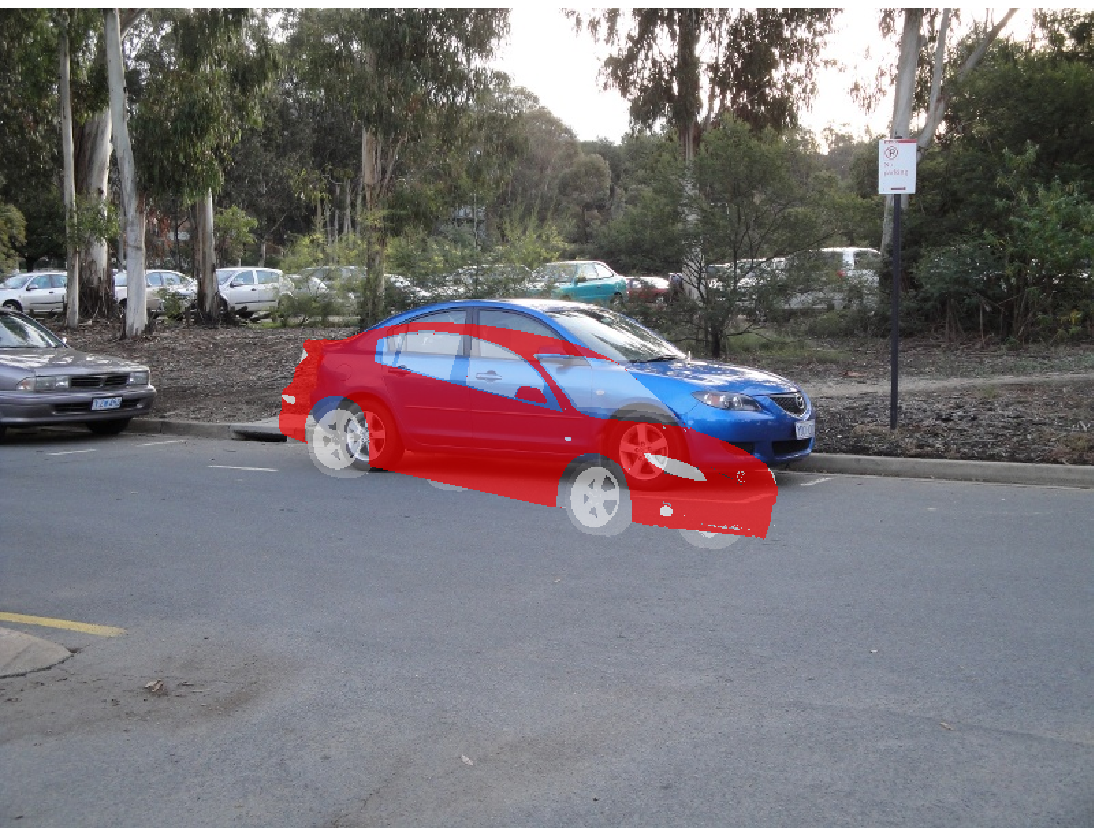}}
\subfigure[Estimated pose]{\label{fig:realphotoEstimatedPose}
   \includegraphics[width=0.48\textwidth]{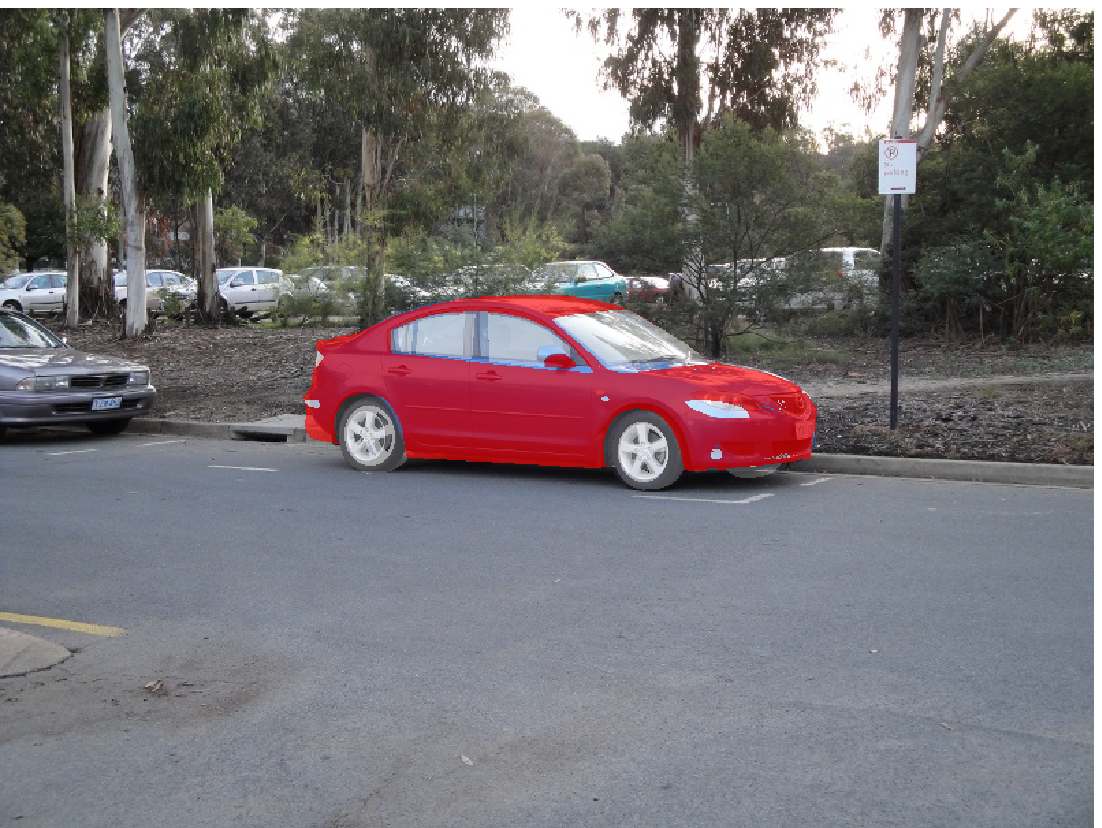}}
\caption[Pose estimation results]{{\bf Pose estimation results.} We
show a sample pose estimation result for a real photograph of a 2005
Mazda 3 car. The car in the original photograph is blue. The
estimated pose is shown by projecting the 3D car model over the
photograph. The 3D car model is in red.}
\label{fig:CombinedResults}
\end{figure}

An example of a correctly estimated pose for a
starting pose within a 16\% deviation from the manually matched pose
is shown in Figures \ref{fig:realphotoInitialPose} and
\ref{fig:realphotoEstimatedPose}. Given that we lack an absolute
ground truth estimate, pose estimates were labelled as correct or
incorrect based on their visual similarity to the input image.

\section{Technical Aspects}\label{secTech}

In this section we describe some of the technical  aspects of
the proposed work. The initial code  was implemented in MATLAB
\cite{matlab}, however, components were gradually ported to C
in order to improve performance.

\paradot{3D rendering}
In order to calculate the loss values described in
Section~\ref{secLossFunction}, it was required to render the
surface normals and brightness of a 3D model at a given pose.
Initially, the rendering was done using \emph{model3D}
\cite{model3d}, a BSD licensed MATLAB \cite{matlab} class. As
this rendering was not fast enough for our application, a
separate module was written in C to render the model off-screen
using OpenGL \cite{opengl} pBuffer extension and GLX. This C
module was  used with the MATLAB code using the \emph{MEX}
gateway. Initially, only the rendering was done in C. The
rendered 2D intensity and surface normal matrices were returned
back to MATLAB using the MEX gateway. This seemed to exhaust
memory during the reliability tests described in
Section~\ref{secOptim}. Therefore, the rendering and the loss
calculation were also implemented in C, with only the loss
value returned to MATLAB for use in optimisation.
\begin{table}
  \caption{Rendering and loss calculation times.}
  \label{tblRenderLosscalcTime}
  \begin{center}
  \begin{tabular}{|l|c|r|}\hline
  \bf Approach   & \bf Loss calc.& \bf Render\\ \hline
  MATLAB     & 0.16 s     & 2.28 s \\ \hline
  C/OpenGL   & 0.04 s     & 0.17 s \\ \hline
\end{tabular}
\end{center}
\end{table}
This second approach improved performance in terms of speed and memory
usage. A summary of the time taken to render the image and to
calculate the loss using these approaches are presented in
Table \ref{tblRenderLosscalcTime}.

\paradot{3D models}
Triangulated 3D car models of significant complexity and detail
in the AutoDesk 3DS file format were used for the work in this
paper. These models were purchased from online 3D model vendors
and had in the order of 30,000 nodes and in the order of 50,000
triangles.

\paradot{Running times}
A typical Downhill Simplex minimisation required in the order
of 100--200 loss function evaluations. Using the C based loss
calculation and OpenGL rendering, pose estimation in artificial
images took around  1 minute for models with more than 30,000
nodes. Recent work done in \cite{moreno2008pose} on pose
estimation using point correspondences, takes more than 3
minutes (200 seconds) for an artificial image of a model with
only 80 points. Hence, despite being a pixel based method, the
performance of our approach is very encouraging.

\paradot{Possible improvements}
The OpenGL context needs to be initialised each time the loss
is calculated, when the C module (MEX) to calculate the loss is
invoked from MATLAB. A further speed-up could be obtained by
implementing the entire code (rendering and optimisation) in C,
whereby the time spent on initialising the OpenGL context could
be saved, as this needs to be done only once. It was also noted
that although hardware accelerated OpenGL performs fast
rendering, reading the rendered pixels back to main memory
causes a performance bottleneck. The loss calculation may also
be done in the graphics hardware itself, using GLSL or GPU
computing, in order to avoid this bottleneck.

\section{Discussion}\label{secDisc}

\paradot{Summary}
A method to register a known 3D model on a given 2D image is
presented in this paper. The correlation between attributes in
the 2D image and projected 3D model are analysed in order to
arrive at a correct pose estimate. The method differs from
existing 2D-3D registration methods found in the literature.
The proposed method requires only a single view of the object.
It does not require a motion sequence and works on a static
image from a given view. Also, the method does not require the
camera parameters to be known a priori. Explicit point
correspondences or matched features (which are hard to obtain
when comparing 3D models and image modalities) need not be
known beforehand. The method can recover the full 3D pose of an
object. It does not require prior training or learning. As the
method can handle 3D models of high complexity and detail, it
could be used for applications that require detailed analysis
of 2D images. It is particularly useful in situations where a
known 3D model is used as a ground truth for analysing a 2D
photograph. The method has been currently tested on real and
artificial photographs of cars with promising results.

\paradot{Outlook}
A planned application of the method is to analyse images of
damaged cars. A known 3D model of the damaged car will be
registered on the image to be analysed, using the proposed
registration method. This will be used as a ground truth. The
method could be extended further to simultaneously identify the
type of the car while estimating its pose, by optimising the
loss function for a number of 3D models and selecting the model
with the lowest loss value. More sophisticated optimisation
methods may be used to improve results further.

\paradot{Conclusion}
We conclude from our results that the linearly invariant loss
function derived in Section~\ref{secLossFunction} can be used
to estimate the pose of cars from real photographs. We also
demonstrate that the \emph{Downhill Simplex} method can be
effectively used to optimise the loss function in order to
obtain the correct pose. Allowing simplex re-initialisations
makes the method more robust against local minima. The
possibility of needing such re-initialisations can be
significantly reduced by clipping the background of the image
when calculating the loss. Despite being a direct pixel based
method (as opposed to a feature/point based method), the
performance of our method is very encouraging in comparison
with other recent approaches, as discussed in
Section~\ref{secTech}.


\begin{small}

\end{small}

\end{document}